\documentclass[a4paper]{article}

\usepackage{INTERSPEECH2021}
\usepackage{comment} 
\usepackage{diagbox} 
\usepackage{slashbox} 
\usepackage{multirow} 
\usepackage{hyperref} 
\usepackage{upgreek}
\usepackage{stmaryrd}
\title{SUPERB: Speech processing Universal PERformance Benchmark}
\name{Shu-wen Yang$^1$, Po-Han Chi\thanks{$^*$Equal contribution; sorted alphabetically}$^{1 *}$, Yung-Sung Chuang$^{1 *}$, Cheng-I Jeff Lai$^{2 *}$, Kushal Lakhotia$^{3 *}$, \\Yist Y. Lin$^{1 *}$, Andy T. Liu$^{1 *}$, Jiatong Shi$^{4 *}$,
Xuankai Chang$^6$, Guan-Ting Lin$^1$, \\Tzu-Hsien Huang$^1$, Wei-Cheng Tseng$^1$, Ko-tik Lee$^1$, Da-Rong Liu$^1$, Zili Huang$^4$, Shuyan Dong$^{5 \dagger}$, Shang-Wen Li\thanks{$^{\dagger}$Work done independently outside Amazon employment}$^{5 \dagger}$, Shinji Watanabe$^6$, Abdelrahman Mohamed$^3$, Hung-yi Lee$^1$}
\address{
  $^1$National Taiwan University, Taiwan\\
  $^2$Massachusetts Institute of Technology, USA\\
  $^3$Facebook AI Research, USA\\
  $^4$Johns Hopkins University, USA\\
  $^5$Amazon AI, USA\\
  $^6$Carnegie Mellon University, USA
}
\email{leo19941227@gmail.com, kushall@fb.com, clai24@mit.edu, jshi34@jhu.edu, shangwel@amazon.com, shinjiw@ieee.org, hungyilee@ntu.edu.tw}

\begin{document}
\maketitle

\begin{abstract}
Self-supervised learning (SSL) has proven vital for advancing research in natural language processing (NLP) and computer vision (CV).
The paradigm pretrains \textit{a shared model} on large volumes of unlabeled data and achieves state-of-the-art (SOTA) \textit{for various tasks with minimal adaptation}.
However, the speech processing community lacks a similar setup to systematically explore the paradigm.
To bridge this gap, we introduce Speech processing Universal PERformance Benchmark (SUPERB).
SUPERB is a leaderboard to benchmark the performance of a shared model across a wide range of speech processing tasks with minimal architecture changes and labeled data.
Among multiple usages of the shared model, we especially focus on extracting the representation learned from SSL for its preferable re-usability.
We present a simple framework to solve SUPERB tasks by learning task-specialized \textit{lightweight} prediction heads on top of the \textit{frozen shared} model.
Our results demonstrate that the framework is promising as SSL representations show competitive generalizability and accessibility across SUPERB tasks.
We release SUPERB as a challenge with a leaderboard\footnote{\label{leaderboard}https://superbbenchmark.org: SUPERB welcomes pretrained model submissions. The framework described in this paper will be used in the constrained track, in which the pretrained models are frozen, and the prediction heads for downstream tasks are the same for all pretrained models. We will open an unconstrained track for submissions with any approach, including finetuning pretrained models and other non-SSL approaches in the future.} and a benchmark toolkit\footnote{\label{toolkit}https://github.com/s3prl/s3prl: All the materials are open-sourced and reproducible in s3prl toolkit which supports to benchmark most existing and customized pretrained models.} to fuel the research in representation learning and general speech processing.
\end{abstract}


\section{Introduction}

Starting from ELMo~\cite{elmo} and BERT~\cite{bert} in NLP, the effectiveness of SSL is evident in various domains~\cite{donotstoppretraining, visual_pretraining}.
It is becoming a new principle to solve problems by pretraining a shared model with self-supervision tasks on a large amount of unlabeled data to encode general-purpose knowledge.
The model can then be specialized in various downstream tasks through concatenating prediction layers and simple finetuning.
This approach achieves SOTA performance in many applications.

SSL is desirable for its outstanding performance as well as generalizability and re-usability across tasks to democratize deep learning to various application scenarios. 
Developing deep neural networks is expensive nowadays in terms of data collection, modeling, computing power, and training time.
Repeating the same process for each specific use case is time- and cost- prohibitive for both academic and industrial researchers.
SSL can significantly speed up and lower the entry barrier for model development, as the pretrained model is powerful to encode generally applicable knowledge, and only requires low resources to extract task-specific knowledge for different use cases.
Well-established benchmark, such as GLUE~\cite{glue} in NLP and VISSL~\cite{vissl} in CV, is essential to evaluate pretrained models' generalizability and re-usability across a wide range of tasks.

SSL has been explored in speech, including pretraining with generative loss~\cite{apc1, mockingjay,tera,decoar2}, discriminative loss~\cite{cpc, wav2vec,vq_wav2vec,wav2vec2}, or multi-task~\cite{pase,pase+}.
Researchers have investigated these SSL models' capabilities on tasks including phoneme classification~\cite{cpc,apc1}, speaker identification~\cite{apc1,mockingjay}, speaker verification~\cite{sv-wav2vec2}, emotion recognition~\cite{pase}, ASR~\cite{tera,wav2vec,decoar2,pase+}, speech translation~\cite{apc1}, spoken language understanding~\cite{lai2020semi}, voice conversion~\cite{lin2020fragmentvc} and TTS~\cite{ssl-for-tts}.
While these works showed promising results of SSL on various speech processing tasks, unlike CV or NLP areas, they were investigated with different datasets and experimental setups.
Absence of a shared benchmark makes it hard to compare and draw insights across the techniques.
Furthermore, existing works explored a limited number of tasks or require heavyweight downstream training~\cite{tera,wav2vec,wav2vec2}, blurring the generalizability and re-usability of SSL models across tasks.
Both factors limit the impact of SSL on speech processing in research and industry.

We introduce Speech processing Universal PERformance Benchmark (SUPERB) to address the problem.
SUPERB aims to 360-degree examine models' capability and collects various tasks with limited labeled data from speech communities to align with common research interests.
There are existing benchmarks proposed to evaluate representations extracted from SSL pretrained models~\cite{zerospeech,non-semantic}.
\cite{zerospeech} focuses on representations' quality without any downstream training, and~\cite{non-semantic} excludes the content recognition tasks like ASR.
Compared to existing efforts, SUPERB targets at the direct usability of pretrained models on various popular tasks through any usage\footnote{\label{multi-approaches}Finetuning pretrained models or using them as representation extractors are two common usages.}.
As finetuning pretrained models typically requires huge resources and hinders the re-usability, in this paper, we focus on investigating a simple framework solving all SUPERB tasks with a \textit{frozen}, \textit{shared} pretrained model, and \textit{lightweight} prediction heads finetuned for each task.
Our results show that the framework yields competitive performance compared to traditional supervised pipelines by leveraging powerful SSL representations, and they outperform log mel filterbank (FBANK), a widely used feature in all speech domains, by a large margin.
Both results demonstrate the possibility of developing powerful, generalizable, and reusable pretrained models to democratize the advance in speech processing.
We invite researchers to participate and submit new results to drive the research frontier together\textsuperscript{\ref{leaderboard}}.

\section{Speech processing Universal PERformance Benchmark}

We establish and release Speech processing Universal PERformance Benchmark (SUPERB), aiming to offer the community a standard and comprehensive testbed for evaluating the generalizability of pretrained models on various tasks covering all aspects of speech.
General speech processing can be categorized into discriminative and generative tasks.
The former discriminates from continuous speech into discrete decisions like \textit{a match} in query-by-example, \textit{words} in ASR, and \textit{classes} in speaker identification; the latter generates continuous speech from any input like TTS, voice conversion, and source separation.
We focus on the former for the initial release of SUPERB\footnote{SUPERB is a long-term maintained and continuously developing project. More pretrained models will be included in the leaderboard, and we plan to release generative tasks as the second challenge, like voice conversion and source separation.}.
Tasks are designed with the following principles: (1) conventional evaluation protocols from speech communities, (2) publicly available datasets for everyone to participate, (3) limited labeled data to effectively benchmark the generalizability of models.
Ten tasks are presented here to investigate four aspects of speech: content, speaker, semantics, and paralinguistics.

\subsection{Content}

Four tasks are collected from ASR and Spoken Term Detection communities.
The former aims to \textit{transcribe} speech into text content; the latter is to \textit{detect} the spoken content with minimal effort even without transcribing.

\textbf{Phoneme Recognition, PR} transcribes an utterance into the smallest content units.
We include alignment modeling in the PR task to avoid the potential inaccurate forced alignment.
LibriSpeech~\cite{librispeech} train-clean-100/dev-clean/test-clean subsets are adopted in SUPERB for training/validation/testing.
Phoneme transcriptions are obtained from the LibriSpeech official \textit{g2p-model-5} and the conversion script in Kaldi \textit{librispeech s5} recipe.
The evaluation metric is phone error rate (PER).

\textbf{Automatic Speech Recognition, ASR} transcribes utterances into words.
While PR analyzes the improvement in modeling phonetics, ASR reflects the significance of the improvement in a real-world scenario.
LibriSpeech train-clean-100/dev-clean/test-clean subsets are used for training/validation/testing.
The evaluation metric is word error rate (WER).

\textbf{Keyword Spotting, KS} detects preregistered keywords by classifying utterances into a predefined set of words.
The task is usually performed on-device for the fast response time.
Thus, accuracy, model size, and inference time are all crucial.
We choose the widely used Speech Commands dataset v1.0~\cite{speech_commands} for the task.
The dataset consists of ten classes of keywords, a class for silence, and an \textit{unknown} class to include the false positive.
The evaluation metric is accuracy (ACC).

\textbf{Query by Example Spoken Term Detection, QbE} detects a spoken term (query) in an audio database (documents) by binary discriminating a given pair of query and document into a match or not.
The English subset in QUESST 2014~\cite{quesst2014} challenge is adopted since we focus on investigating English as the first step.
The evaluation metric is maximum term weighted value (MTWV) which balances misses and false alarms.

\subsection{Speaker}

Three tasks are collected to analyze speaker modeling.

\textbf{Speaker Identification, SID} classifies each utterance for its speaker identity as a multi-class classification, where speakers are in the same predefined set for both training and testing.
The widely used VoxCeleb1~\cite{voxceleb1} is adopted, and the evaluation metric is accuracy (ACC).

\textbf{Automatic Speaker Verification, ASV} verifies whether the speakers of a pair of utterances match as a binary classification, and speakers in the testing set may not appear in the training set.
Thus, ASV is more challenging than SID.
VoxCeleb1~\cite{voxceleb1} is used without VoxCeleb2 training data and noise augmentation.
The evaluation metric is equal error rate (EER).

\textbf{Speaker Diarization, SD} predicts \textit{who is speaking when} for each timestamp, and multiple speakers can speak simultaneously.
The model has to encode rich speaker characteristics for each frame and should be able to represent mixtures of signals.
LibriMix~\cite{cosentino2020librimix} is adopted where LibriSpeech train-clean-100/dev-clean/test-clean are used to generate mixtures for training/validation/testing.
We focus on the two-speaker scenario as the first step.
The time-coded speaker labels were generated using alignments from Kaldi LibriSpeech ASR model.
The evaluation metric is diarization error rate (DER).

\subsection{Semantics}

Two tasks are collected from Spoken Language Understanding (SLU) community.
While most works for these tasks are done in two stages: transcribing speech into text and predicting semantics on transcribed text, we focus on inferring high-level semantics directly from raw audio in an end-to-end fashion.

\textbf{Intent Classification, IC} classifies utterances into predefined classes to determine the intent of speakers.
We use the Fluent Speech Commands~\cite{lugosch2019speech} dataset, where each utterance is tagged with three intent labels: action, object, and location.
The evaluation metric is accuracy (ACC).

\textbf{Slot Filling, SF} predicts a sequence of semantic slot-types from an utterance, like a slot-type \textit{FromLocation} for a spoken word \textit{Taipei}, which is known as a slot-value.
Both slot-types and slot-values are essential for an SLU system to function~\cite{lai2020semi}.
The evaluation metrics thus include slot-type F1 score and slot-value CER~\cite{tomashenko2019recent}.
Audio SNIPS~\cite{lai2020semi} is adopted, which synthesized multi-speaker utterances for SNIPS~\cite{coucke2018snips}.
Following the standard split in SNIPS, US-accent speakers are further selected for training, and others are for validation/testing.

\subsection{Paralinguistics}

\textbf{Emotion Recognition, ER} predicts an emotion class for each utterance.
The most widely used ER dataset IEMOCAP~\cite{iemocap} is adopted, and we follow the conventional evaluation protocol: we drop the unbalance emotion classes to leave the final four classes (neutral, happy, sad, angry) with a similar amount of data points and cross-validates on five folds of the standard splits.
The evaluation metric is accuracy (ACC).

\section{Framework: Universal Representation}

Our framework aims to explore \textit{how simple and general the solution can be}. 
Thus, we freeze the parameters of pretrained models across tasks and extract fixed representations to be fed into each task-specialized prediction head (small downstream model).
Compared to previous setups in speech representation learning~\cite{tera,wav2vec,vq_wav2vec}, the framework puts an explicit constraint on downstream models to be as lightweight as possible for all tasks, as their parameter size and required training resources are also crucial for the framework to be simple and re-usable in various use cases.
With the above principles, the pretrained model solving all SUPERB tasks in this framework would be a universal representation encoder.
In the following, we first describe the SSL pretrained models leveraged and then introduce the downstream models and training policies.


\subsection{Self-supervised pretrained models}

SSL models explored in this paper are summarized in Table~\ref{table:upstreams} and categorized into three learning approaches: generative modeling, discriminative modeling, and multi-task learning.

\textbf{Generative modeling} has long been a prevailing approach to learn speech representation~\cite{apc1,mockingjay,decoar2}.
Instances of generative modeling investigated here include APC~\cite{apc1}, VQ-APC~\cite{vq_apc}, Mockingjay~\cite{mockingjay}, TERA~\cite{tera}, NPC~\cite{npc} and DeCoAR 2.0~\cite{decoar2}.
APC adopts the language model-like pretraining scheme on a sequence of acoustic features (FBANK) with unidirectional RNN and generates future frames conditioning on past frames.
VQ-APC further applies vector-quantization (VQ) layers onto APC's representation to make it compact and low bit-rate.
Mockingjay adopts the BERT-like pretraining on Transformer encoders by masking the input acoustic features in time axis and re-generating the masked parts.
TERA extends Mockingjay to further mask the frequency bins.
NPC improves the inference speed upon APC by replacing RNN with CNN and changing the future generation to masked reconstruction as Mockingjay.
DeCoAR 2.0 improves upon Mockingjay by inserting a VQ layer right before the final prediction like VQ-APC, and is trained by larger input mask, larger batch size, and more unlabeled data.

\textbf{Discriminative modeling} for SSL studied here include CPC~\cite{cpc,modified_cpc}, wav2vec~\cite{wav2vec}, vq-wav2vec~\cite{vq_wav2vec}, wav2vec 2.0~\cite{wav2vec2} and HuBERT~\cite{hsu2021hubert}.
CPC discriminates the correlated positive samples from negative samples with contrastive InfoNCE loss, which maximizes the mutual information between raw data and representations.
Modified CPC~\cite{modified_cpc} and wav2vec~\cite{wav2vec} proposed several architecture changes to improve CPC.
vq-wav2vec introduces a VQ module to wav2vec. The module discretizes speech into a sequence of tokens after InfoNCE pretraining.
Tokens are used as pseudo-text to train a BERT as did in NLP for contextualized representations.
wav2vec 2.0 merges the pipeline of vq-wav2vec into one end-to-end training scheme by applying time masking in the latent space and replacing BERT's token prediction with InfoNCE's negative sampling to handle the intractable normalization on continuous speech.
Motivated by DeepCluster~\cite{caron18deepcluster}, HuBERT~\cite{hsu2021hubert} enables BERT's token prediction via off-line clustering on representations.
The clustered labels at the masked locations are then predicted.

\textbf{Multi-task learning} is applied in PASE+~\cite{pase+}, where lots of pretraining objectives are adopted: waveform generation, prosody features regression, contrastive InfoMax objectives, and more.
Multiple contaminations are also applied to input speech like reverberation and additive noise.

\subsection{Downstream models and policies}

We design our framework to keep the downstream models and their finetuning simple, while ensuring the performance across pretrained models is comparable and the best model in each task is competitive.
Since the last-layer representation is not always the best, the framework collects multiple hidden states from the pretrained model and weighted-sum them as the final representation.
For a fair comparison, we also limit the space for downstream hyper-parameters tuning\footnote{We search for the best learning rate across 1e-1 to 1e-7 in log-scale for each combination of SSL representation and the downstream tasks. More details about the allowed hyper-parameters tuning will be available as we announce the challenge, but there will not be many hyper-parameters to keep tuning simple.}.
Downstream models and algorithms are summarized in the following and will be released in detail as a part of the challenge policy.

\textbf{PR}, \textbf{KS}, \textbf{SID}, \textbf{IC}, \textbf{ER} are simple tasks that are solvable with linear downstream models.
Hence, we use a frame-wise linear transformation for PR with CTC loss; mean-pooling followed by a linear transformation with cross-entropy loss for utterance-level tasks (KS, SID, IC, and ER).
These five tasks also serve as the direct indication of representations' quality following the conventional linear evaluation protocol.

For \textbf{ASR}, a vanilla 2-layer 1024-unit BLSTM is adopted and optimized by CTC loss on characters. The trained model is decoded with LibriSpeech official 4-gram LM powered by KenLM~\cite{kenlm} and flashlight~\cite{wav2letter++} toolkit.
We mostly follow the system proposed by GTTS-EHU for QUESST at MediaEval 2014 \cite{gttsehu} for \textbf{QbE} but replace the conventional supervised phoneme posteriorgram (PPG) with SSL representations.
We run Dynamic Time Warping\cite{dtw} on all hidden states separately with standard distance functions and obtain a score for each query-document pair.
The best distance function / hidden state pair is reported.
Regarding \textbf{SF}, slot-type labels are represented as special tokens to wrap the slot-values in transcriptions.
SF is then re-formulated as an ASR problem.
The finetuning scheme is the same as in our ASR task, except for the pre-processing to encode slot-types into transcriptions and post-processing to decode slot-types and slot-values from hypotheses.
As for \textbf{ASV}, we adopt the well-known x-vector~\cite{snyder2018x} as the downstream model and change Softmax loss to AMSoftmax loss with the same hyper-parameters as \cite{voxceleb1}.
The simple cosine-similarity backend is used to produce pairwise matching scores.
We employ the end-to-end training scheme with permutation-invariant training (PIT) loss \cite{fujita2019end} to \textbf{SD}, instead of using clustering-based methods. We leverage a single-layer 512-unit LSTM for the downstream model.
\begin{table*}[ht]
\centering
\setlength{\tabcolsep}{3pt}
\caption{
Details of investigated SSL representations. LibriSpeech and LibriLight are denoted as LS and LL, respectively. For the pretraining methods, we abbreviate "vector quantization" as VQ, "future" as F, "masked" as M, "generation" as G, "contrastive discrimination" as C, and "token prediction/classification" as P. Parameters for both pretraining and inference are counted.
}
\resizebox{0.98\textwidth}{!}{
\begin{tabular}{|l||c|c|c|c|c|c|c|}
\hline
Method & Network & \#Params & Stride & Input & Corpus & Pretraining & Official Github \\ \hline \hline

FBANK & - & 0 & 10ms & waveform & - & - & - \\ \hline

PASE+~\cite{pase+} & SincNet, 7-Conv, 1-QRNN & 7.83M & 10ms & waveform & LS 50 hr & multi-task & santi-pdp / pase \\ \hline

APC~\cite{apc1} & 3-GRU & 4.11M & 10ms & FBANK & LS 360 hr & F-G & iamyuanchung / APC \\

VQ-APC~\cite{vq_apc} & 3-GRU & 4.63M & 10ms & FBANK & LS 360 hr & F-G + VQ & iamyuanchung / VQ-APC \\

NPC~\cite{npc} & 4-Conv, 4-Masked Conv & 19.38M & 10ms & FBANK & LS 360 hr & M-G + VQ & Alexander-H-Liu / NPC \\

Mockingjay~\cite{mockingjay} & 12-Trans & 85.12M & 10ms & FBANK & LS 360 hr & time M-G & s3prl / s3prl \\

TERA~\cite{tera} & 3-Trans & 21.33M & 10ms & FBANK & LS 960 hr & time/freq M-G & s3prl / s3prl \\ 

DeCoAR 2.0~\cite{decoar2} & 12-Trans & 89.84M & 10ms & FBANK & LS 960 hr & time M-G + VQ & awslabs / speech-representations \\ 
\hline

modified CPC~\cite{modified_cpc} & 5-Conv, 1-LSTM & 1.84M & 10ms & waveform & LL 60k hr & F-C & facebookresearch / CPC\_audio \\

wav2vec~\cite{wav2vec} & 19-Conv & 32.54M & 10ms & waveform & LS 960 hr & F-C & pytorch / fairseq \\

vq-wav2vec~\cite{vq_wav2vec} & 20-Conv & 34.15M & 10ms & waveform & LS 960 hr & F-C + VQ & pytorch / fairseq \\

wav2vec 2.0 Base~\cite{wav2vec2} & 7-Conv 12-Trans & 95.04M & 20ms & waveform & LS 960 hr & M-C + VQ & pytorch / fairseq \\

wav2vec 2.0 Large~\cite{wav2vec2} & 7-Conv 24-Trans & 317.38M & 20ms & waveform & LL 60k hr & M-C + VQ & pytorch / fairseq \\

HuBERT Base~\cite{hsu2021hubert} & 7-Conv 12-Trans & 94.68M & 20ms & waveform & LS 960 hr & M-P + VQ & pytorch / fairseq \\

HuBERT Large~\cite{hsu2021hubert} & 7-Conv 24-Trans & 316.61M & 20ms & waveform & LL 60k hr & M-P + VQ & pytorch / fairseq \\
\hline
\end{tabular}}
\label{table:upstreams}
\end{table*}

\begin{table*}[ht]
\centering
\caption{Evaluating various SSL representations on various downstream tasks. The numbers are collected with public-available checkpoints or codes, and we welcome researchers to re-submit the results to our online leaderboard.
}
\resizebox{1.0\textwidth}{!}{
\begin{tabular}{|l||r|r|r|r|r||rr|r|rr|r|r|}
\hline
\multirow{2}{*}{} & PR & KS & IC & SID & ER & \multicolumn{2}{c|}{ASR (WER)} & QbE & \multicolumn{2}{c|}{SF} & ASV & SD \\ \cline{2-13}

& PER $\downarrow$ & Acc $\uparrow$ & Acc $\uparrow$ & Acc $\uparrow$ & Acc $\uparrow$ & w/o $\downarrow$ & w/ LM $\downarrow$ & MTWV $\uparrow$ & F1 $\uparrow$ & CER $\downarrow$ & EER $\downarrow$ & DER $\downarrow$ \\ \hline \hline

FBANK & 82.01 & 8.63 & 9.10 & 8.5E-4 & 35.39 & 23.18 & 15.21 & 0.0058 & 69.64 & 52.94 & 9.56 & 10.05 \\ \hline

PASE+~\cite{pase+} & 58.87 & 82.54 & 29.82 & 37.99 & 57.86 & 25.11 & 16.62 & 0.0072 & 62.14 & 60.17 & 11.61 & 8.68 \\ \hline

APC~\cite{apc1} & 41.98 & 91.01 & 74.69 & 60.42 & 59.33 & 21.28 & 14.74 & 0.0310 & 70.46 & 50.89 & 8.56 & 10.53 \\

VQ-APC~\cite{vq_apc} & 41.08 & 91.11 & 74.48 & 60.15 & 59.66 & 21.20 & 15.21 & 0.0251 & 68.53 & 52.91 & 8.72 & 10.45 \\

NPC~\cite{npc} & 43.81 & 88.96 & 69.44 & 55.92 & 59.08 & 20.20 & 13.91 & 0.0246 & 72.79 & 48.44 & 9.4 & 9.34 \\

Mockingjay~\cite{mockingjay} & 70.19 & 83.67 & 34.33 & 32.29 & 50.28 & 22.82 & 15.48 & 6.6E-04 & 61.59 & 58.89 & 11.66 & 10.54 \\

TERA~\cite{tera} & 49.17 & 89.48 & 58.42 & 57.57 & 56.27 & 18.17 & 12.16 & 0.0013 & 67.50 & 54.17 & 15.89 & 9.96 \\

DeCoAR 2.0~\cite{decoar2} & 14.93 & 94.48 & 90.80 & 74.42 & 62.47 & 13.02 & 9.07 & 0.0406 & 83.28 & 34.73 & 7.16 & 6.59 \\

\hline

modified CPC~\cite{modified_cpc} & 42.54 & 91.88 & 64.09 & 39.63 & 60.96 & 20.18 & 13.53 & 0.0326 & 71.19 & 49.91 & 12.86 & 10.38 \\

wav2vec~\cite{wav2vec} & 31.58 & 95.59 & 84.92 & 56.56 & 59.79 & 15.86 & 11.00 & 0.0485 & 76.37 & 43.71 & 7.99 & 9.9 \\

vq-wav2vec~\cite{vq_wav2vec} & 33.48 & 93.38 & 85.68 & 38.80 & 58.24 & 17.71 & 12.80 & 0.0410 & 77.68 & 41.54 & 10.38 & 9.93 \\

wav2vec 2.0 Base~\cite{wav2vec2} & 5.74 & 96.23 & 92.35 & 75.18 & 63.43 & 6.43 & 4.79 & 0.0233 & 88.30 & 24.77 & 6.02 & 6.08 \\

wav2vec 2.0 Large~\cite{wav2vec2} & 4.75 & \textbf{96.66} & 95.28 & 86.14 & 65.64 & 3.75 & 3.10 & 0.0489 & 87.11 & 27.31 & 5.65 &  \textbf{5.62} \\

HuBERT Base~\cite{hsu2021hubert} & 5.41 & 96.30 & 98.34 & 81.42 & 64.92 & 6.42 & 4.79 & \textbf{0.0736} & 88.53 & 25.20 & \textbf{5.11} & 5.88 \\

HuBERT Large~\cite{hsu2021hubert} & \textbf{3.53} & 95.29 & \textbf{98.76} & \textbf{90.33} & \textbf{67.62} & \textbf{3.62} & \textbf{2.94} & 0.0353 & \textbf{89.81} & \textbf{21.76} & 5.98 & 5.75 \\
\hline
\end{tabular}
}
\label{table:exp}
\vspace*{-4mm}
\end{table*}

\section{Experiment}

To extract representations from pretrained models, we follow the official release as summarized in Table~\ref{table:upstreams} for model definitions, pretrained weights, and extraction pipelines if not mentioning specifically.
Some noteworthy details are:
(1) NPC repository is used to pretrain APC and VQ-APC as it is more flexible\footnote{It is preferable for its on-the-fly FBANK extraction to enable testing representations on more corpora. Its APC implementation is mostly the same as the official but with CMVN on FBANK. Its VQ-APC is an improved version as stated in the official repository.}.
(2) For vq-wav2vec, we do not propagate through BERT since its BERT implementation limits the utterance length which is not long enough for some tasks.

We present the results in Table~\ref{table:exp}.
For the tasks using linear models, FBANK cannot work on any task, while SSL representations all perform well to some degree with different specializations.
It is a surprise that wav2vec 2.0 and HuBERT conquers PR and IC with just linear models and outperforms others by a large margin.
Their results on SID and ER are also highly competitive.
FBANK achieves competitive performance when allowing non-linear downstream models in ASR, SF, ASV, and SD, and yields better performance than some SSL representations.
We also observe that the ranking on PR aligns with ASR weakly, while a significant improvement on phonetics still transfers to ASR, like wav2vec, wav2vec 2.0, and HuBERT.
Furthermore, wav2vec 2.0 and HuBERT demonstrate that it becomes much easier than before to train an ASR system by leveraging powerful SSL representations.
HuBERT ranks the top one in QbE with MTWV 0.074.
The prevailing feature for QbE is PPG which we implemented with TIMIT due to the current focus on English, and the result of TIMIT PPG is 0.052 in MTWV, suggesting that HuBERT turns out to be a very competitive representation for QbE.
As for SF, we can also observe a significant improvement from wav2vec 2.0 and HuBERT over all other representations.
The CER in SF is generally high compared to ASR as many slot-values are named entities.
The results in ASV and SD show that many SSL representations are worse than FBANK when it comes to real-world speaker problems beyond SID, while HuBERT improves upon popular FBANK from 9.56 to 5.10 without additional VoxCeleb2 or augmentation.
Although we find it non-trivial for SSL representations to generalize to all SUPERB tasks, wav2vec 2.0 and HuBERT achieve highly competitive performance with only lightweight prediction heads trainable, compared to traditional supervised techniques. The experiment results exhibit the efficacy of developing a more generalizable and re-usable pretrained model.

\section{Conclusion}

We present SUPERB, a challenge to generally benchmark the capability of SSL pretrained models on speech processing.
We demonstrate a simple framework to solve all SUPERB tasks which leverages a frozen, shared pretrained model and achieves competitive performance with minimal architecture changes and downstream finetuning.
We have open-sourced the evaluation toolkit\textsuperscript{\ref{toolkit}} and will release the detailed challenge policy on the leaderboard website\textsuperscript{\ref{leaderboard}}.
We welcome the community to participate and drive the research frontier.

\bibliographystyle{IEEEtran}
\bibliography{mybib}


\end{document}